%% file: SLT.tex
\documentclass[10pt, conference]{IEEEtran}
\IEEEoverridecommandlockouts
\usepackage{cite}
\usepackage{amsmath,amssymb,amsfonts}
\usepackage{algorithmic}
\usepackage{graphicx}
\usepackage{textcomp}
\usepackage{xcolor}
\newcommand{\red}[1]{\textcolor{red}{#1}}
\usepackage{pifont}
\usepackage{booktabs}
\newcommand{\std}[1]{\,{\scriptsize\color{gray}$\pm$#1}}
\def\BibTeX{{\rm B\kern-.05em{\sc i\kern-.025em b}\kern-.08em
    T\kern-.1667em\lower.7ex\hbox{E}\kern-.125emX}}

\begin{document}

\title{AMRD: Adaptive Multi-Teacher Relational Distillation for Lightweight Speech Emotion Recognition
}

\author{
    \IEEEauthorblockN{
        Yuqi Li\IEEEauthorrefmark{1}\textsuperscript{\dag}, 
        Yi-Cheng Lin\IEEEauthorrefmark{2}\textsuperscript{*\dag}, 
        Xianglong Wang\IEEEauthorrefmark{3}, 
        Kuo Yang\IEEEauthorrefmark{3}, 
        Xiaoqin Feng\IEEEauthorrefmark{3}, \\ 
        Yixuan Wang\IEEEauthorrefmark{4}, 
        Huiran Duan\IEEEauthorrefmark{1}, 
        Yingli Tian\IEEEauthorrefmark{1}\textsuperscript{*}
    }
    \vspace{0.15cm}
    \IEEEauthorblockA{\IEEEauthorrefmark{1}Department of Electrical Engineering, The City College of New York}
    \IEEEauthorblockA{\IEEEauthorrefmark{2}National Taiwan University}
    \IEEEauthorblockA{\IEEEauthorrefmark{3}Wyze Inc.}
    \IEEEauthorblockA{\IEEEauthorrefmark{4}University of Minnesota - Twin Cities}
    \IEEEauthorblockA{\textsuperscript{\dag}Equal contribution \qquad \textsuperscript{*}Corresponding authors}
}

\maketitle

\begin{abstract}
On-device speech emotion recognition (SER) is critical for real-time applications, yet large self-supervised models that excel at SER are too costly for edge devices. Multi-teacher knowledge distillation can compress them into a lightweight student, but two challenges remain: teacher reliability varies across batches, and logit-level distillation ignores inter-sample relational structure. We propose Adaptive Multi-teacher Relational Distillation (AMRD) to address both. A one-class SVM on each teacher's logit similarity matrix assigns per-batch weights favoring more coherent teachers. A relational distillation loss aligns teacher and student similarity matrices, capturing structure that logit matching misses. On IEMOCAP and CREMA-D datasets across four student architectures, AMRD outperforms single-teacher distillation baselines in most settings, and ablations confirm both components yield complementary gains.
\end{abstract}


\section{Introduction}
\label{sec:intro}

Speech emotion recognition (SER) aims to identify emotional states from speech signals and underpins applications such as human-computer interaction, mental health monitoring, and call center analytics \cite{koolagudi2012emotion}. The task remains challenging because emotion annotation is inherently subjective \cite{8682170, fang25b_interspeech} and speaker variability in prosody and speaking style hinders cross-speaker generalization. Because no single acoustic representation captures all such variations equally well, combining multiple complementary models is a promising direction \cite{huang2025mifuse}.

Among such models, recent self-supervised learning (SSL) approaches such as WavLM \cite{chen2022wavlm} and data2vec \cite{baevski2022data2vec} have substantially improved SER \cite{wu2024emosuperbindepthlookspeech}, yet even their base variants contain ${\sim}$94\,M parameters, making edge deployment impractical. Knowledge distillation (KD) \cite{hinton2015distilling} addresses this by training a compact student to mimic a large teacher's output distribution. When multiple SSL teachers are available, multi-teacher KD (MTKD) can distill their complementary knowledge into a single student \cite{fukuda17_interspeech, 11434750}; this is attractive for SER because different SSL pre-training objectives capture different aspects of the speech signal \cite{yang21c_interspeech}.

However, applying MTKD to SER introduces two challenges that existing methods do not adequately address. 
The first is \emph{teacher reliability heterogeneity}. When teachers differ in architecture and pre-training objective, their prediction quality varies across samples. Moreover, this variation is not static: it shifts from batch to batch as the data conditions (speaker identity, emotional intensity, recording environment) change \cite{xia2018instance, 9745778}. 
Standard MTKD uses uniform or fixed aggregation weights, which cannot adapt to this per-batch fluctuation and may propagate unreliable soft targets to the student. 
The second challenge is \emph{incomplete knowledge transfer}. 
Conventional logit-level KD transfers only the predicted class distribution for each sample independently, ignoring the relational structure among samples in the teacher's feature space. 
In SER, emotionally similar utterances form natural clusters in feature space \cite{10552082}; preserving this structure provides a training signal that sample-wise logit matching misses.

We propose \textbf{AMRD} (\textbf{A}daptive \textbf{M}ulti-teacher \textbf{R}elational \textbf{D}istillation), an MTKD framework for SER that addresses both challenges. For teacher aggregation, AMRD employs a one-class SVM \cite{scholkopf2001estimating} to score each teacher's per-batch reliability from its pairwise logit structure, so that teachers whose predictions are more coherent receive higher aggregation weights. This adaptive weighting is particularly important in SER, where emotion subjectivity and speaker diversity cause teacher reliability to shift across batches. To complement logit-level distillation, AMRD introduces relational similarity matrix distillation (RSMD), which minimizes the mean squared error between pairwise cosine similarity matrices computed from teacher and student representations, encouraging the student to preserve each teacher's inter-sample similarity structure. Since SVM weighting and RSMD are used only during training, neither strategy incurs additional cost at inference time.

\begin{figure*}[t]
  \centering
  \includegraphics[width=0.88\linewidth]{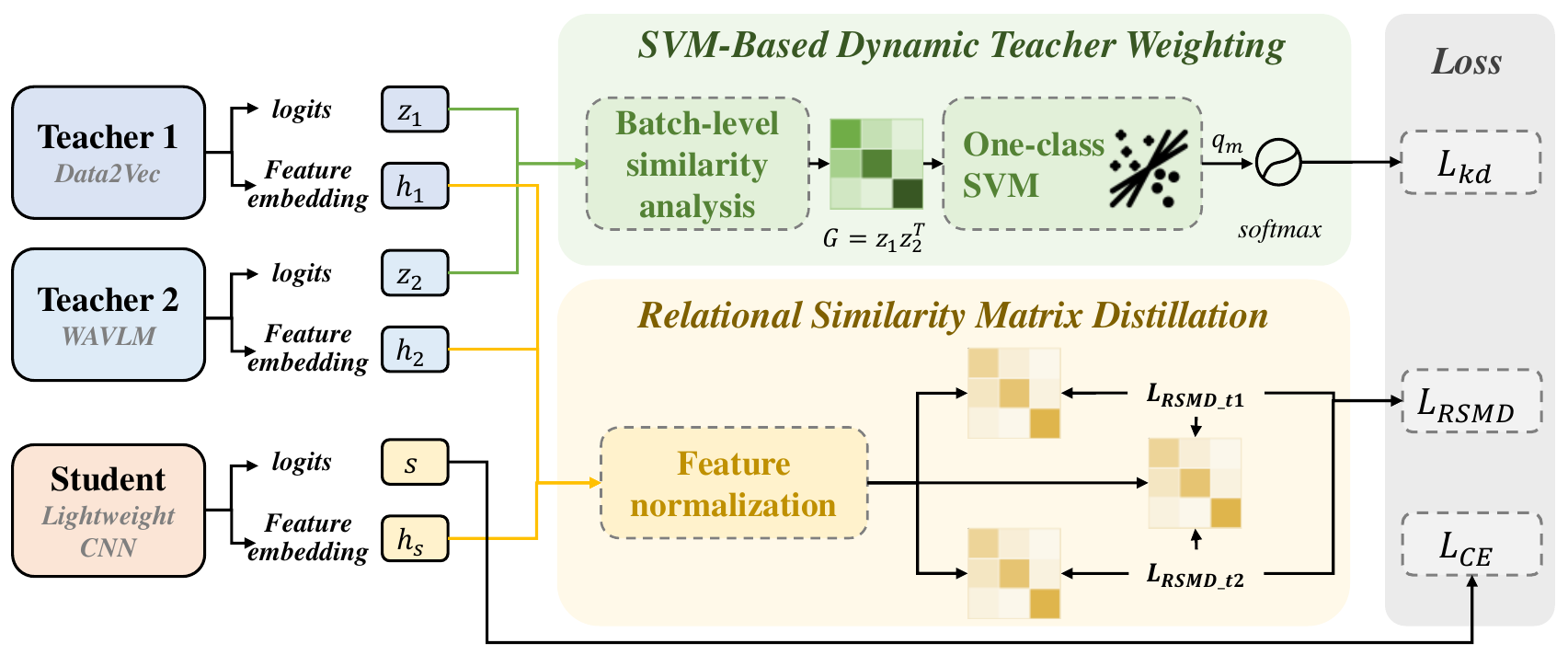}

  \caption{Overview of AMRD. Two frozen SSL teachers (data2vec Large and WavLM Base+) produce logits and features for each batch. A one-class SVM is fitted on each teacher's pairwise cosine similarity matrix of batch logits to obtain per-batch aggregation weights $w_m$; the weighted soft targets guide logit-level distillation. Simultaneously, pairwise cosine similarity matrices (RSMs) are computed from teacher and student features, and the RSMD loss aligns them. Both mechanisms operate only during training.}
  \label{fig:framework}
\end{figure*}

Our contributions are threefold: 
\begin{itemize}
    \item We apply SVM-based dynamic teacher weighting to SER, showing that per-batch adaptive aggregation is effective when teacher reliability varies across data conditions. 
    \item We introduce RSMD, a feature-level relational loss that transfers inter-sample similarity structure from multiple teachers to the student. 
    \item Experiments on IEMOCAP and CREMA-D show that AMRD matches or exceeds the best single-teacher baseline in seven of eight student--dataset settings across four architectures, with gains of up to 2.9\% accuracy.
\end{itemize}

\section{Related Work}
\label{sec:related}
 
\subsection{Feature-based and relational KD.}
At the logit level, DKD \cite{zhao2022dkd} and CKD \cite{li2023ckd} refine soft-target transfer but do not capture inter-sample structure.
FitNets \cite{romero2015fitnets} introduced point-wise intermediate feature matching, which also ignores relational structure. Several methods do transfer relational information: SPKD \cite{tung2019similarity} matches activation Gram matrices, RKD \cite{park2019rkd} penalizes distance and angle discrepancies among sample groups, PKT \cite{passalis2018learning} matches pairwise-similarity distributions via KL divergence, and CCKD \cite{peng2019correlation} aligns correlation structure via RBF kernels. 
All four distill from a single teacher and provide no mechanism to normalize relational targets across feature spaces that differ in scale and distribution. 
CRD~\cite{tian2020contrastive} uses contrastive mutual-information maximization to capture relational knowledge, but requires a memory buffer spanning the entire training set whose entries grow stale as the student updates. 
Zhou et al.~\cite{zhou2024rethinking} use Centered Kernel Alignment (CKA)~\cite{pmlr-v97-kornblith19a}, a normalized variant of the Hilbert-Schmidt Independence Criterion (HSIC)~\cite{gretton2005hsic}, as a distillation loss, but normalizing the kernel matrices by their Frobenius norms removes the matrix-magnitude term that can serve as collapse prevention in an unnormalized formulation (Section~\ref{ssec:RSMD}). 
None of these methods combines relational transfer with adaptive multi-teacher weighting.
 
\subsection{Multi-teacher KD for SER.}
Multi-teacher KD aggregates complementary knowledge from several teachers \cite{fukuda17_interspeech, 11434750}, but most methods use uniform or fixed weights that cannot adapt to per-batch reliability shifts. 
In SER, Bijoy et al.\ \cite{bijoy2025multilingual} proposed cosine-similarity routing for multilingual emotion recognition, where each teacher's per-sample weight is the cosine similarity between its logits and the student's. 
Because this selects the teacher the student already agrees with, it can reinforce existing predictions rather than prioritizing the most reliable source. 
Furthermore, the teacher and student share the same architecture (${\sim}$95\,M parameters) with no compression, and the transfer is logit-level only.
AMRD assesses teacher quality from the teacher's own batch-level logit structure, independent of the student, and combines logit-level distillation with relational transfer under heterogeneous-architecture compression.

\section{Proposed Method}
\label{sec:method}

As illustrated in Fig.~\ref{fig:framework}, AMRD distills knowledge from a set of $M$ frozen SSL teachers into a lightweight student with trainable parameters~$\theta$. We use $M{=}2$ teachers in this work, data2vec \cite{baevski2022data2vec} and WavLM \cite{chen2022wavlm}. Given a batch of $B$ utterances with ground-truth labels $y$, each teacher takes raw waveforms as input and produce, for each utterance, a classification logit vector $\mathbf{z}_m \in \mathbb{R}^C$ over $C$ emotion classes from the final layer and a feature vector $\mathbf{h}_m \in \mathbb{R}^D$ from the layer before the classifier. Likewise, the student takes log-Mel spectrograms and yields logits $\mathbf{s} \in \mathbb{R}^C$ and features $\mathbf{h}_s \in \mathbb{R}^{D'}$, where $D' \neq D$ because the architectures differ. The logits are used for soft-target distillation, and the features for relational distillation. 

\subsection{SVM-Based Dynamic Teacher Weighting}
\label{ssec:svm}

As noted in Section~\ref{sec:intro}, teacher reliability shifts from batch to batch, so the aggregation weights must be adaptive. 
One natural approach is to use scalar confidence measures such as prediction entropy or softmax margin.
However, these per-sample statistics do not capture the joint structure across the batch: a teacher may assign high confidence to individual samples while producing inconsistent logit patterns across samples of the same emotion. 
In SER, emotion classes overlap in acoustic space, so a teacher can produce confident yet mutually inconsistent predictions within the same batch, a failure mode that per-sample metrics such as entropy cannot detect. 
We instead assess each teacher through the pairwise structure of its batch logits. For teacher $m$, we $\ell_2$-normalize its batch logits to obtain $\bar{\mathbf{Z}}_m \in \mathbb{R}^{B \times C}$ and compute a cosine similarity matrix:
\begin{equation}
    \mathbf{G}_m = \bar{\mathbf{Z}}_m \bar{\mathbf{Z}}_m^\top \in \mathbb{R}^{B \times B}.
\label{eq:gram}
\end{equation}
To distill this $B{\times}B$ matrix into a scalar quality score, we fit a one-class SVM with $\mathbf{G}_m$ as a precomputed kernel matrix. 
A one-class SVM learns a compact boundary enclosing the majority of data points; the mean signed distance to this boundary then serves as the teacher's quality score:
\begin{equation}
    q_m = \frac{1}{B}\sum_{i=1}^{B} d_m^{(i)},
\label{eq:svm_score}
\end{equation}
where $d_m^{(i)}$ is the decision value for sample $i$. Intuitively, when a teacher produces coherent logits, the rows of $\mathbf{G}_m$ form a tight cluster and most samples fall inside the SVM boundary, yielding a high $q_m$. 
Scattered or inconsistent predictions push samples outside the boundary and lower $q_m$. 
The one-class SVM hyperparameter $\nu$ controls the upper bound on the fraction of outliers; we fix $\nu{=}0.1$ in all experiments. The aggregation weights are obtained via softmax:
\begin{equation}
    w_m = \frac{\exp(q_m)}{\sum_{j=1}^{M} \exp(q_j)}.
\label{eq:svm_weight}
\end{equation}
Each teacher's logit vector $\mathbf{z}_m$ is softened with a shared temperature $T$, and the aggregated soft-target distribution is:
\begin{equation}
    \mathbf{p}_T = \sum_{m=1}^{M} w_m \, \mathrm{softmax}(\mathbf{z}_m / T).
\label{eq:agg}
\end{equation}
Because the kernel matrix is only $B{\times}B$ with $B{=}16$, the SVM fitting adds negligible overhead to training and is not needed at inference time.

\input{tables/main_iemocap}
\input{tables/main_cremad}

\subsection{Relational Similarity Matrix Distillation}
\label{ssec:RSMD}

Logit-level KD transfers per-sample class distributions but does not preserve how samples relate to each other in the feature space. We address this with an RSMD loss. For each batch, RSMD computes a pairwise cosine similarity matrix, which we call the relational similarity matrix (RSM), from both teacher and student features, and minimizes their difference. 
Unlike RKD \cite{park2019rkd}, which transfers Euclidean distances and angles between embeddings, cosine similarity is invariant to the absolute scale of feature vectors, making it suitable when teacher and student produce features of different magnitudes.

We now detail the RSM construction. Because the teacher and student feature spaces differ in both dimensionality ($D \neq D'$) and distribution, directly comparing feature vectors is not possible. The RSM circumvents this mismatch: by computing pairwise similarities in a $B{\times}B$ matrix, it captures which samples are close to each other regardless of the underlying feature dimension. To ensure that the similarity values are comparable despite differences in feature magnitude and distribution, we apply $z$-score normalization followed by $\ell_2$-normalization to each feature matrix. The $z$-score step independently centers and scales each sample's feature vector to zero mean and unit variance across its $D$ dimensions, removing per-sample magnitude bias; the subsequent $\ell_2$-normalization ensures that the resulting cosine similarities lie in $[-1, 1]$. Given a normalized feature matrix $\hat{\mathbf{H}} \in \mathbb{R}^{B \times D}$, the RSM is:
\begin{equation}
    \mathbf{R} = \hat{\mathbf{H}} \, \hat{\mathbf{H}}^\top \in \mathbb{R}^{B \times B}.
\label{eq:rsm}
\end{equation}
The RSMD loss aligns the student's RSM with each teacher's:
\begin{equation}
    \mathcal{L}_{\text{RSMD}} = \frac{\alpha_{\text{geom}}}{M} \sum_{m=1}^{M} \mathrm{MSE}\bigl(\mathbf{R}_s,\, \mathbf{R}_{t_m}\bigr),
\label{eq:RSMD}
\end{equation}
where $\mathbf{R}_s$ and $\mathbf{R}_{t_m}$ are the student and teacher-$m$ RSMs, $\alpha_{\text{geom}}$ is a scalar weight, and MSE is the element-wise mean squared error. Expanding the Frobenius norm of each per-teacher term:
\begin{equation}
\|\mathbf{R}_s - \mathbf{R}_{t_m}\|_F^2
= \underbrace{\|\mathbf{R}_s\|_F^2}_{\textnormal{regularization}}- 2\underbrace{\|\hat{\mathbf{H}}_{t_m}^\top \hat{\mathbf{H}}_s\|_F^2}_{\textnormal{alignment}}
+\underbrace{\|\mathbf{R}_{t_m}\|_F^2}_{\textnormal{const}},
\label{eq:decomp}
\end{equation}
where $\|\hat{\mathbf{H}}_{t_m}^\top \hat{\mathbf{H}}_s\|_F^2 = \mathrm{tr}(\mathbf{R}_s \mathbf{R}_{t_m})$. Since teacher parameters are frozen, the last term is constant with respect to~$\theta$. Minimizing the MSE therefore simultaneously maximizes $\mathrm{tr}(\mathbf{R}_s \mathbf{R}_{t_m})$, which rewards the student for reproducing the teacher's pairwise similarity pattern, and minimizes $\|\mathbf{R}_s\|_F^2$, which penalizes uniformly high pairwise similarities among student representations.

The $z$-score step ensures that these pairwise similarities are Pearson correlations ($\rho$) rather than cosine similarities: because $z$-scoring gives every row the same squared norm $D{-}1$ (by the definition of sample variance), $\ell_2$-normalization divides all rows by $\sqrt{D{-}1}$, and each RSM entry reduces to $R_{ij} = \rho(\mathbf{h}_i, \mathbf{h}_j)$. Unlike cosine similarity, Pearson correlation is invariant to per-sample affine transformations ($\rho(\alpha\mathbf{h}_i {+} c\mathbf{1},\; \beta\mathbf{h}_j {+} d\mathbf{1}) = \rho(\mathbf{h}_i, \mathbf{h}_j)$ for $\alpha,\beta > 0$), filtering out distributional differences between teacher and student feature spaces. Because $\rho(\mathbf{h}_i, \mathbf{h}_i) {=} 1$, the diagonal of $\mathbf{R}_s {-} \mathbf{R}_t$ is zero and the MSE depends only on off-diagonal correlations.

We average over teachers with uniform weights rather than reusing the SVM-based weights $w_m$ from Section~\ref{ssec:svm}. The SVM scores measure logit-level coherence, which does not necessarily reflect the quality of a teacher's feature-level relational structure: a teacher may produce scattered logits yet still maintain informative inter-sample relationships in its hidden representations. Empirically, uniform averaging yields stable results across all settings, and decoupling the two weighting mechanisms simplifies the overall framework.

\subsection{Training Objective}
\label{ssec:objective}

The overall loss combines cross-entropy on hard labels, KL divergence on aggregated soft targets, and RSMD:
\begin{equation}
    \mathcal{L} = \alpha_{\text{ce}} \, \mathcal{L}_{\text{CE}}(\mathbf{s}, y)
              + \alpha_{\text{kd}} \, T^2 \, \mathrm{KL}(\log \sigma(\mathbf{s}/T) \ ||\ \mathbf{p}_T)
              + \mathcal{L}_{\text{RSMD}},
\label{eq:total}
\end{equation}
where $\sigma$ denotes softmax and $T^2$ compensates for gradient magnitude reduction from temperature scaling \cite{hinton2015distilling}. The coefficients $\alpha_{\text{ce}}$ and $\alpha_{\text{kd}}$ remain fixed throughout training, as the SVM-based weights already provide per-batch adaptivity at the soft-target level. Only the student parameters $\theta$ are updated; all teachers remain frozen. The SVM fitting step is non-differentiable, so the aggregation weights $w_m$ are treated as fixed constants during backpropagation.

\section{Experiments}
\label{sec:exp}

\subsection{Setup}
\label{ssec:setup}

\noindent\textbf{Datasets.}
We evaluate on two widely used SER benchmarks. \textbf{IEMOCAP} \cite{busso2008iemocap} contains approximately 12 hours of audiovisual recordings from 10 speakers across 5 sessions. Following prior work \cite{yang21c_interspeech, 10887615}, we select four emotions (angry, happy, neutral, sad) and merge \emph{excited} into \emph{happy}. We adopt a leave-one-session-out 5-fold evaluation \cite{wu2024emosuperbindepthlookspeech}, where each session serves as the test set in turn. \textbf{CREMA-D} \cite{cao2014cremad} consists of 7,442 clips from 91 actors. We use all 6 emotion classes with 5-fold cross-validation.

\input{tables/ablation}

\noindent\textbf{Models.}
We build two teacher SER models by appending a single linear classifier to data2vec Large \cite{baevski2022data2vec} and WavLM Base+ \cite{chen2022wavlm}, following previous approaches~\cite{yang21c_interspeech, lin24i_interspeech}. Both are fully fine-tuned per fold; the best checkpoint is selected by validation unweighted accuracy (UA). We evaluate four student architectures spanning a wide parameter range, all taking 64-dim log-Mel spectrograms as input: LightSERNet+ (LSP+, 0.78\,M) \cite{lightsernet}, MobileNetV3 (MNv3, 1.52\,M) \cite{howard2019mobilenetv3}, EfficientNet-B0 (EB0, 4.01\,M) \cite{tan2019efficientnet}, and ResNet18 (R18, 11.7\,M) \cite{he2016resnet}.

\noindent\textbf{Training.}
We use AdamW \cite{loshchilov2018decoupled} with learning rate $10^{-4}$, weight decay $10^{-4}$, and cosine annealing for 50 epochs. The batch size is 16. Distillation hyperparameters are $T{=}4$, $\alpha_{\text{ce}}{=}1$, $\alpha_{\text{kd}}{=}1$, $\alpha_{\text{geom}}{=}0.1$, and SVM $\nu{=}0.1$. Mixed-precision training and gradient clipping (max norm 1.0) are applied. Model selection is based on validation UA.

\noindent\textbf{Baselines.}
We compare AMRD against the following: (1)~\textbf{No KD}, where the student is trained with cross-entropy only; (2)~\textbf{Vanilla KD} \cite{hinton2015distilling}; (3)~\textbf{DKD} \cite{zhao2022dkd}, which decouples target-class and non-target-class distillation; (4)~\textbf{CKD} \cite{li2023ckd}, which averages KL divergence over multiple temperatures; (5)~\textbf{RKD} \cite{park2019rkd}, which uses pairwise distance and angle losses. All baselines use the same two teachers, student architecture, and training configuration for a fair comparison. All baseline-specific hyperparameters use the default values from their original papers. We report weighted accuracy (WA) and UA averaged over all folds. WA reflects overall sample-level performance, while UA averages per-class accuracy and is therefore more informative when the class distribution is imbalanced.

\subsection{Main Results}
\label{ssec:results}

Tables~\ref{tab:iemocap} and~\ref{tab:cremad} report WA and UA on IEMOCAP and CREMA-D, respectively. Single-teacher results use either WavLM or data2vec as the teacher; AMRD uses both.

On IEMOCAP, AMRD achieves the highest WA and UA across all four students. The largest gain appears for MNv3, where AMRD reaches 47.21\% UA, 2.9 points above the best single-teacher result (WavLM+DKD, 44.27\%). R18 and EB0 improve by 1.2 and 1.1 UA points over their respective best single-teacher baselines (WavLM+CKD and D2V+CKD). For LSP+, AMRD matches the best single-teacher UA (52.30\% vs.\ 52.26\% for D2V+RKD) while improving WA by 1.1 points (50.30\%$\to$51.41\%). 

On CREMA-D, AMRD leads on three of four students. LSP+ reaches 56.37\% UA, exceeding the best single-teacher result (D2V+KD, 54.65\%) by 1.7 points; R18 and EB0 also achieve the highest WA and UA, though by smaller margins. The exception is MNv3, where all methods cluster within a narrow 1.7-point UA range (34.82--36.56\%) and most KD variants, including AMRD (36.09\%), fall below the no-distillation baseline (36.39\%). This indicates that distillation provides limited benefit for this student--dataset pair, regardless of the teacher configuration.

The two datasets exhibit different teacher dynamics. On IEMOCAP, data2vec yields the stronger single-teacher distillation for LSP+ (best UA: 52.26 vs.\ 51.00 for WavLM), while WavLM is more effective for MNv3 (44.27 vs.\ 43.07); this variation across students highlights the value of adaptive teacher weighting. On CREMA-D, data2vec generally leads as the stronger teacher, yet AMRD still surpasses the best D2V-based result for three of four students, indicating that the secondary teacher contributes a useful signal even when it is not the dominant source. Overall, AMRD matches or exceeds the better single-teacher result in seven of eight student--dataset settings without requiring prior knowledge of which teacher suits each student.

From a deployment perspective, LSP+ contains only 0.78\,M parameters, over 120$\times$ fewer than the smaller teacher WavLM Base+ (${\sim}$94\,M). With AMRD, this student achieves 52.30\% UA on IEMOCAP and 56.37\% on CREMA-D, improving over the no-distillation baseline by 2.8 and 3.4 points, respectively, demonstrating that the framework transfers meaningful knowledge even under extreme compression.

\subsection{Ablation Study}
\label{ssec:ablation}
Table~\ref{tab:ablation} isolates the contribution of each AMRD component by toggling SVM-based weighting and RSMD independently. We report LSP+ and EB0 as representatives of the smallest and mid-capacity students. Each component individually improves over the uniform-weight, no-RSMD baseline in all settings. On IEMOCAP, SVM weighting alone raises UA by 0.4--0.7 points; RSMD alone adds 0.2--0.3 points. On CREMA-D, both effects are larger: SVM boosts EB0 UA by 1.8 points (39.75\%$\to$41.51\%) and LSP+ by 1.2 points; RSMD alone yields 0.6--0.7 points. Combining both achieves the best result in three of four settings. The exception is LSP+ on IEMOCAP, where SVM-only reaches 52.45\% UA, but adding RSMD lowers this to 52.30\%. We attribute this to LSP+'s very limited capacity (0.78\,M), which makes it sensitive to the relational constraint when emotion boundaries are ambiguous, as in IEMOCAP's conversational speech.

\red{\subsection{Hyperparameter Sensitivity}}
\label{ssec:sensitivity}
 
\input{tables/sensitivity}

We analyze the sensitivity of AMRD to the relational loss weight $\alpha_{\text{geom}}$ under the same teacher pair, using LSP+ and EB0 as representative students. As shown in Table~\ref{tab:sensitivity}, performance is stable for moderate values ($\alpha_{\text{geom}} \in [0.05, 0.2]$). Removing the relational term ($\alpha_{\text{geom}}{=}0$) reduces UA in three of four settings, confirming that RSMD provides complementary supervision beyond logit-level distillation. LSP+ on IEMOCAP slightly favors $\alpha_{\text{geom}}{=}0.05$, consistent with the ablation finding that this smallest student is more sensitive to the relational constraint on IEMOCAP. When $\alpha_{\text{geom}}$ is increased to $0.5$, performance drops in all settings, indicating that an overly strong relational loss over-regularizes the student. We also varied the distillation temperature $T \in \{2, 4, 6, 8\}$ and the one-class SVM parameter $\nu \in \{0.05, 0.1, 0.2\}$; results remained stable around the defaults $T{=}4$ and $\nu{=}0.1$. We therefore use $\alpha_{\text{geom}}{=}0.1$, $T{=}4$, and $\nu{=}0.1$ in all main experiments.

\section{Conclusion}
We presented AMRD, a multi-teacher knowledge distillation framework for lightweight SER. AMRD uses SVM-based dynamic weighting to adapt teacher aggregation per batch and introduces RSMD, a relational loss that aligns pairwise similarity structure between teacher and student feature spaces. Both mechanisms operate only during training and add no inference cost. On IEMOCAP and CREMA-D, AMRD matches or exceeds the better single-teacher baseline in seven of eight student--dataset settings. The smallest student, LSP+, with over 120$\times$ fewer parameters than WavLM Base+, achieves 52.30\% and 56.37\% UA on the two benchmarks, improving over the no-distillation baseline by 2.8 and 3.4 points, demonstrating that adaptive multi-teacher distillation can enable on-device emotion recognition under extreme compression.

Future work includes extending AMRD to more than two teachers, incorporating noise-robust training for real-world deployment, and evaluating cross-corpus transfer. Since knowledge distillation can amplify or suppress biases present in teacher models~\cite{lin24b_interspeech}, systematic fairness evaluation across speaker demographics is also an important step before deployment~\cite {lin24i_interspeech, lin2026fairspeech}.

\section{Limitations}

\noindent\textbf{Number of teachers.}
All experiments use two teachers. Although both the SVM weighting and RSMD are defined for arbitrary $M$, scaling to more teachers introduces additional hyperparameter choices (teacher selection, per-teacher loss weighting) and increases training cost linearly in $M$. Validating AMRD with a larger and more diverse teacher pool is left for future work.
 
\noindent\textbf{Audio-only modality.}
AMRD processes speech only. Emotion is also conveyed through facial expression and linguistic content, and fusing modalities can resolve ambiguities irrecoverable from audio alone. Extending the framework to multimodal inputs requires aligned multi-channel corpora with per-modality teachers, which constitutes a different research problem.
 
\noindent\textbf{In-corpus evaluation.}
All experiments use an in-corpus setting where training and test data come from the same dataset. Cross-corpus and cross-language evaluations would provide stronger evidence of generalization but require careful handling of label-set and recording-condition mismatches, which we leave for future work.

\section{Generative AI Use Disclosure}

Generative AI tools assisted in the linguistic polishing of the manuscript. 
The authors remain solely responsible for the research design, experiments, analysis, and reported results. 
AI tools did not contribute to the substantive scientific content.

\newpage
\bibliographystyle{IEEEtran}
\bibliography{mybib}

\end{document}

%% file: tables/main_iemocap.tex
\begin{table*}[t]
\caption{\textbf{IEMOCAP results.} WA and UA (\%) for single-teacher and multi-teacher distillation across four student architectures. \textbf{Bold}: best per column; \textit{italic}: second best. Standard deviations across 5 folds in {\tiny\color{gray}gray}.}
\label{tab:iemocap}
\centering
\setlength{\tabcolsep}{6pt}
\begin{tabular}{@{}ll cc cc cc cc@{}}
\toprule
 &  & \multicolumn{2}{c}{LSP+} & \multicolumn{2}{c}{MNv3} & \multicolumn{2}{c}{R18} & \multicolumn{2}{c}{EB0} \\
\cmidrule(lr){3-4} \cmidrule(lr){5-6} \cmidrule(lr){7-8} \cmidrule(lr){9-10}
Teacher & Method & WA & UA & WA & UA & WA & UA & WA & UA \\
\midrule
\multicolumn{2}{@{}l}{\textit{Teacher (upper bound)}} \\
WavLM & -- & \multicolumn{8}{c}{WA: 58.93\std{3.2} \quad UA: 59.86\std{1.9}} \\
D2V   & -- & \multicolumn{8}{c}{WA: 64.76\std{4.6} \quad UA: 65.19\std{5.4}} \\
\midrule
\multicolumn{2}{@{}l}{\textit{No distillation}} \\
-- & Baseline & 48.60\std{4.1} & 49.53\std{2.8} & 41.13\std{3.2} & 41.46\std{2.8} & 47.40\std{3.5} & 50.58\std{2.8} & 46.56\std{2.6} & 45.67\std{4.8} \\
\midrule
\multicolumn{2}{@{}l}{\textit{Single-teacher KD (WavLM)}} \\
WavLM & KD  & 49.55\std{2.2} & 50.54\std{4.3} & 44.21\std{2.8} & 43.86\std{3.1} & 46.94\std{3.6} & 50.01\std{3.5} & 47.29\std{2.2} & 46.85\std{2.7} \\
WavLM & DKD & 50.14\std{3.3} & 50.08\std{4.2} & \textit{45.06}\std{1.8} & \textit{44.27}\std{2.1} & 49.09\std{3.3} & 50.82\std{3.6} & 46.95\std{2.1} & 46.54\std{3.4} \\
WavLM & CKD & 49.25\std{2.3} & 50.28\std{3.4} & 44.05\std{1.7} & 43.65\std{2.3} & \textit{51.13}\std{2.8} & \textit{51.45}\std{2.0} & \textit{47.52}\std{2.2} & 47.21\std{3.8} \\
WavLM & RKD & \textit{50.30}\std{3.6} & 51.00\std{2.9} & 44.11\std{1.7} & 43.52\std{2.2} & 48.67\std{3.5} & 49.57\std{3.6} & 45.89\std{1.8} & 46.49\std{3.4} \\
\midrule
\multicolumn{2}{@{}l}{\textit{Single-teacher KD (D2V)}} \\
D2V & KD  & 48.69\std{2.5} & 50.01\std{4.3} & 43.02\std{1.2} & 42.62\std{2.3} & 50.12\std{1.2} & 50.91\std{2.4} & 45.85\std{1.6} & 46.54\std{4.1} \\
D2V & DKD & 48.59\std{3.8} & 50.52\std{3.6} & 43.20\std{2.2} & 43.07\std{2.2} & 49.93\std{1.5} & 51.22\std{3.5} & 46.00\std{2.1} & 46.34\std{4.7} \\
D2V & CKD & 49.23\std{2.7} & 51.44\std{3.8} & 42.77\std{2.2} & 42.47\std{2.9} & 50.38\std{1.6} & 50.56\std{3.1} & 46.35\std{1.8} & \textit{47.37}\std{2.2} \\
D2V & RKD & 48.31\std{0.9} & \textit{52.26}\std{0.7} & 43.15\std{1.7} & 42.75\std{2.3} & 47.46\std{2.0} & 49.70\std{3.3} & 45.18\std{3.0} & 46.81\std{3.8} \\
\midrule
\multicolumn{2}{@{}l}{\textit{Multi-teacher (ours)}} \\
Both & \textbf{AMRD} & \textbf{51.41}\std{1.1} & \textbf{52.30}\std{2.3} & \textbf{46.27}\std{2.1} & \textbf{47.21}\std{3.4} & \textbf{51.31}\std{0.4} & \textbf{52.65}\std{1.6} & \textbf{47.97}\std{1.7} & \textbf{48.49}\std{2.6} \\
\bottomrule
\end{tabular}
\end{table*}

%% file: tables/main_cremad.tex
\begin{table*}[t]
\caption{\textbf{CREMA-D results.} WA and UA (\%) for single-teacher and multi-teacher distillation across four student architectures. \textbf{Bold}: best per column; \textit{italic}: second best. Standard deviations across 5 folds in {\tiny\color{gray}gray}.}
\label{tab:cremad}
\centering
\setlength{\tabcolsep}{6pt}
\begin{tabular}{@{}ll cc cc cc cc@{}}
\toprule
 &  & \multicolumn{2}{c}{LSP+} & \multicolumn{2}{c}{MNv3} & \multicolumn{2}{c}{R18} & \multicolumn{2}{c}{EB0} \\
\cmidrule(lr){3-4} \cmidrule(lr){5-6} \cmidrule(lr){7-8} \cmidrule(lr){9-10}
Teacher & Method & WA & UA & WA & UA & WA & UA & WA & UA \\
\midrule
\multicolumn{2}{@{}l}{\textit{Teacher (upper bound)}} \\
WavLM & -- & \multicolumn{8}{c}{WA: 71.76\std{2.2} \quad UA: 72.02\std{2.1}} \\
D2V   & -- & \multicolumn{8}{c}{WA: 72.43\std{0.5} \quad UA: 72.84\std{0.5}} \\
\midrule
\multicolumn{2}{@{}l}{\textit{No distillation}} \\
-- & Baseline & 52.81\std{1.4} & 52.96\std{1.5} & 36.42\std{3.0} & \textit{36.39}\std{3.0} & 52.73\std{2.2} & 52.87\std{2.3} & 40.38\std{1.7} & 40.47\std{1.6} \\
\midrule
\multicolumn{2}{@{}l}{\textit{Single-teacher KD (WavLM)}} \\
WavLM & KD  & 53.37\std{1.6} & 53.50\std{1.6} & 36.19\std{2.4} & 36.14\std{2.4} & 55.26\std{2.5} & 55.25\std{2.6} & 41.35\std{3.4} & 41.46\std{3.3} \\
WavLM & DKD & 53.15\std{1.4} & 53.39\std{1.2} & 35.46\std{2.2} & 35.30\std{2.2} & 54.87\std{2.5} & 54.97\std{2.7} & 40.76\std{1.6} & 40.73\std{1.5} \\
WavLM & CKD & 54.36\std{1.4} & 54.60\std{1.3} & 36.07\std{2.7} & 36.02\std{2.7} & 54.37\std{2.8} & 54.34\std{2.7} & 40.16\std{3.6} & 40.24\std{3.4} \\
WavLM & RKD & 53.52\std{1.3} & 53.53\std{1.3} & \textit{36.46}\std{2.1} & 36.32\std{2.2} & 54.02\std{2.9} & 54.15\std{2.9} & 39.45\std{2.0} & 39.53\std{1.9} \\
\midrule
\multicolumn{2}{@{}l}{\textit{Single-teacher KD (D2V)}} \\
D2V & KD  & \textit{54.79}\std{2.1} & \textit{54.65}\std{2.2} & 35.32\std{1.8} & 35.20\std{1.7} & \textit{55.60}\std{1.9} & \textit{55.73}\std{1.9} & \textit{41.91}\std{1.9} & \textit{42.11}\std{1.7} \\
D2V & DKD & 53.48\std{2.2} & 53.55\std{2.2} & \textbf{36.56}\std{1.3} & \textbf{36.48}\std{1.2} & 55.15\std{1.5} & 55.34\std{1.5} & 41.74\std{1.7} & 41.79\std{1.7} \\
D2V & CKD & 52.92\std{1.9} & 52.92\std{1.9} & 35.47\std{1.9} & 35.30\std{1.8} & 55.33\std{1.3} & 55.51\std{1.4} & 41.07\std{1.7} & 41.18\std{1.8} \\
D2V & RKD & 52.67\std{1.4} & 52.82\std{1.3} & 34.93\std{1.8} & 34.82\std{1.8} & 52.76\std{2.0} & 52.78\std{2.0} & 39.79\std{1.2} & 39.76\std{1.2} \\
\midrule
\multicolumn{2}{@{}l}{\textit{Multi-teacher (ours)}} \\
Both & \textbf{AMRD} & \textbf{56.27}\std{1.5} & \textbf{56.37}\std{1.4} & 36.14\std{1.2} & 36.09\std{1.1} & \textbf{56.01}\std{2.7} & \textbf{56.16}\std{2.6} & \textbf{42.07}\std{2.7} & \textbf{42.27}\std{2.6} \\
\bottomrule
\end{tabular}
\end{table*}

%% file: tables/ablation.tex
\begin{table*}[t]
\caption{\textbf{Ablation study.} Effect of SVM-based adaptive teacher weighting and RSMD loss on IEMOCAP and CREMA-D (UA and WA, \%). ``Uni'' = uniform teacher weights; ``SVM'' = SVM-based adaptive weights. \textbf{Bold}: best per column; \textit{italic}: second best. Standard deviations across folds in {\tiny\color{gray}gray}.}
\label{tab:ablation}
\centering
\setlength{\tabcolsep}{6pt}
\begin{tabular}{cc cc cc cc cc}
\toprule
 &  & \multicolumn{4}{c}{IEMOCAP} & \multicolumn{4}{c}{CREMA-D} \\
\cmidrule(lr){3-6} \cmidrule(lr){7-10}
 &  & \multicolumn{2}{c}{LSP+} & \multicolumn{2}{c}{EB0} & \multicolumn{2}{c}{LSP+} & \multicolumn{2}{c}{EB0} \\
\cmidrule(lr){3-4} \cmidrule(lr){5-6} \cmidrule(lr){7-8} \cmidrule(lr){9-10}
Weight & RSMD & WA & UA & WA & UA & WA & UA & WA & UA \\
\midrule
Uni & \ding{55} & 51.12\std{1.0} & 52.02\std{2.2} & 46.92\std{1.6} & 47.45\std{2.5} & 54.40\std{1.0} & 54.50\std{1.1} & 39.58\std{2.4} & 39.75\std{2.4} \\
SVM & \ding{55} & \textbf{51.56}\std{1.1} & \textbf{52.45}\std{2.2} & \textit{47.58}\std{1.7} & \textit{48.12}\std{2.6} & \textit{55.60}\std{1.3} & \textit{55.70}\std{1.4} & \textit{41.32}\std{2.6} & \textit{41.51}\std{2.6} \\
Uni & \ding{51} & 51.28\std{1.0} & 52.18\std{2.2} & 47.24\std{1.7} & 47.78\std{2.5} & 55.00\std{1.1} & 55.10\std{1.2} & 40.25\std{2.5} & 40.42\std{2.5} \\
SVM & \ding{51} & \textit{51.41}\std{1.1} & \textit{52.30}\std{2.3} & \textbf{47.97}\std{1.7} & \textbf{48.49}\std{2.6} & \textbf{56.27}\std{1.5} & \textbf{56.37}\std{1.4} & \textbf{42.07}\std{2.7} & \textbf{42.27}\std{2.6} \\
\bottomrule
\end{tabular}
\end{table*}

%% file: tables/sensitivity.tex
\begin{table}[t]
\caption{\textbf{Sensitivity to $\alpha_{\text{geom}}$.} UA (\%) of LightSERNet+ and EfficientNet-B0 under different geometric-mean weights $\alpha_{\text{geom}}$. \textbf{Bold}: best per column. Standard deviations across folds in {\tiny\color{gray}gray}.}
\label{tab:sensitivity}
\centering
\begin{tabular}{c cc cc}
\toprule
 & \multicolumn{2}{c}{IEMOCAP} & \multicolumn{2}{c}{CREMA-D} \\
\cmidrule(lr){2-3} \cmidrule(lr){4-5}
$\alpha_{\text{geom}}$ & LSP+ & EB0 & LSP+ & EB0 \\
\midrule
0.00 & 52.45\std{2.2} & 47.58\std{1.7} & 55.70\std{1.4} & 41.51\std{2.6} \\
0.05 & \textbf{52.62}\std{2.0} & 48.06\std{2.1} & 56.08\std{1.6} & 41.94\std{2.4} \\
0.10 & 52.30\std{2.3} & \textbf{48.49}\std{2.6} & \textbf{56.37}\std{1.4} & \textbf{42.27}\std{2.6} \\
0.20 & 51.98\std{2.5} & 48.21\std{2.2} & 56.11\std{1.7} & 42.06\std{2.8} \\
0.50 & 51.36\std{2.7} & 47.74\std{2.5} & 55.58\std{1.9} & 41.62\std{3.0} \\
\bottomrule
\end{tabular}
\end{table}

%% file: mybib.bib
@misc{wu2024emosuperbindepthlookspeech,
      title={EMO-SUPERB: An In-depth Look at Speech Emotion Recognition}, 
      author={Haibin Wu and Huang-Cheng Chou and Kai-Wei Chang and Lucas Goncalves and Jiawei Du and Jyh-Shing Roger Jang and Chi-Chun Lee and Hung-Yi Lee},
      year={2024},
      eprint={2402.13018},
      archivePrefix={arXiv},
      primaryClass={eess.AS}, 
}

@ARTICLE{cao2014cremad,
  author={Cao, Houwei and Cooper, David G. and Keutmann, Michael K. and Gur, Ruben C. and Nenkova, Ani and Verma, Ragini},
  journal={IEEE Transactions on Affective Computing}, 
  title={CREMA-D: Crowd-Sourced Emotional Multimodal Actors Dataset}, 
  year={2014},
  volume={5},
  number={4},
  pages={377-390},
  doi={10.1109/TAFFC.2014.2336244}
}

@article{busso2008iemocap,
  title={IEMOCAP: Interactive emotional dyadic motion capture database},
  author={Busso, Carlos and Bulut, Murtaza and Lee, Chi-Chun and Kazemzadeh, Abe and Mower, Emily and Kim, Samuel and Chang, Jeannette N and Lee, Sungbok and Narayanan, Shrikanth S},
  journal={Language resources and evaluation},
  volume={42},
  number={4},
  pages={335--359},
  year={2008},
  publisher={Springer}
}

@InProceedings{baevski2022data2vec,
  title = 	 {data2vec: A General Framework for Self-supervised Learning in Speech, Vision and Language},
  author =       {Baevski, Alexei and Hsu, Wei-Ning and Xu, Qiantong and Babu, Arun and Gu, Jiatao and Auli, Michael},
  booktitle = 	 {Proceedings of the 39th International Conference on Machine Learning},
  pages = 	 {1298--1312},
  year = 	 {2022},
  editor = 	 {Chaudhuri, Kamalika and Jegelka, Stefanie and Song, Le and Szepesvari, Csaba and Niu, Gang and Sabato, Sivan},
  volume = 	 {162},
  series = 	 {Proceedings of Machine Learning Research},
  month = 	 {17--23 Jul},
  publisher =    {PMLR}
}

@ARTICLE{chen2022wavlm,
  author={Chen, Sanyuan and Wang, Chengyi and Chen, Zhengyang and Wu, Yu and Liu, Shujie and Chen, Zhuo and Li, Jinyu and Kanda, Naoyuki and Yoshioka, Takuya and Xiao, Xiong and Wu, Jian and Zhou, Long and Ren, Shuo and Qian, Yanmin and Qian, Yao and Wu, Jian and Zeng, Michael and Yu, Xiangzhan and Wei, Furu},
  journal={IEEE Journal of Selected Topics in Signal Processing}, 
  title={WavLM: Large-Scale Self-Supervised Pre-Training for Full Stack Speech Processing}, 
  year={2022},
  volume={16},
  number={6},
  pages={1505-1518}}

@InProceedings{he2016resnet,
author = {He, Kaiming and Zhang, Xiangyu and Ren, Shaoqing and Sun, Jian},
title = {Deep Residual Learning for Image Recognition},
booktitle = {Proceedings of the IEEE Conference on Computer Vision and Pattern Recognition (CVPR)},
year = {2016}
}

@article{hinton2015distilling,
  title={Distilling the Knowledge in a Neural Network},
  author={Hinton, Geoffrey and Vinyals, Oriol and Dean, Jeff},
  journal={stat},
  volume={1050},
  pages={9},
  year={2015}
}

@InProceedings{howard2019mobilenetv3,
author = {Howard, Andrew and Sandler, Mark and Chu, Grace and Chen, Liang-Chieh and Chen, Bo and Tan, Mingxing and Wang, Weijun and Zhu, Yukun and Pang, Ruoming and Vasudevan, Vijay and Le, Quoc V. and Adam, Hartwig},
title = {Searching for MobileNetV3},
booktitle = {Proceedings of the IEEE/CVF International Conference on Computer Vision (ICCV)},
month = {October},
year = {2019}
}

@inproceedings{fukuda17_interspeech,
  title     = {{Efficient Knowledge Distillation from an Ensemble of Teachers}},
  author    = {Takashi Fukuda and Masayuki Suzuki and Gakuto Kurata and Samuel Thomas and Jia Cui and Bhuvana Ramabhadran},
  year      = {2017},
  booktitle = {{Interspeech 2017}},
  pages     = {3697--3701},
  doi       = {10.21437/Interspeech.2017-614}
}

@inproceedings{yang21c_interspeech,
  title     = {{SUPERB: Speech Processing Universal PERformance Benchmark}},
  author    = {Shu-wen Yang and Po-Han Chi and Yung-Sung Chuang and Cheng-I Jeff Lai and Kushal Lakhotia and Yist Y. Lin and Andy T. Liu and Jiatong Shi and Xuankai Chang and Guan-Ting Lin and Tzu-Hsien Huang and Wei-Cheng Tseng and Ko-tik Lee and Da-Rong Liu and Zili Huang and Shuyan Dong and Shang-Wen Li and Shinji Watanabe and Abdelrahman Mohamed and Hung-yi Lee},
  year      = {2021},
  booktitle = {{Interspeech 2021}},
  pages     = {1194--1198},
  issn      = {2958-1796}
}

@misc{huang2025mifuse,
      title={MI-Fuse: Label Fusion for Unsupervised Domain Adaptation with Closed-Source Large-Audio Language Model}, 
      author={Hsiao-Ying Huang and Yi-Cheng Lin and Hung-yi Lee},
      year={2025},
      eprint={2509.20706},
      archivePrefix={arXiv},
      primaryClass={cs.CL}, 
}

@InProceedings{tan2019efficientnet,
  title = 	 {{E}fficient{N}et: Rethinking Model Scaling for Convolutional Neural Networks},
  author =       {Tan, Mingxing and Le, Quoc},
  booktitle = 	 {Proceedings of the 36th International Conference on Machine Learning},
  pages = 	 {6105--6114},
  year = 	 {2019},
  volume = 	 {97},
  series = 	 {Proceedings of Machine Learning Research},
  publisher =    {PMLR}
}

@article{koolagudi2012emotion,
  title={Emotion recognition from speech: a review},
  author={Koolagudi, Shashidhar G and Rao, K Sreenivasa},
  journal={International journal of speech technology},
  volume={15},
  number={2},
  pages={99--117},
  year={2012},
  publisher={Springer}
}

@INPROCEEDINGS{8682170,
  author={Chou, Huang-Cheng and Lee, Chi-Chun},
  booktitle={ICASSP 2019 - 2019 IEEE International Conference on Acoustics, Speech and Signal Processing (ICASSP)}, 
  title={Every Rating Matters: Joint Learning of Subjective Labels and Individual Annotators for Speech Emotion Classification}, 
  year={2019},
  volume={},
  number={},
  pages={5886-5890}
  }

@inproceedings{fang25b_interspeech,
  title     = {{Meta-PerSER: Few-Shot Listener Personalized Speech Emotion Recognition via Meta-learning}},
  author    = {Shi-Xin Fang and Liang-Yeh Shen and Yi-Cheng Lin and Huang-Cheng Chou and Hung-yi Lee},
  year      = {2025},
  booktitle = {{Interspeech 2025}},
  pages     = {136--140}
}

@InProceedings{park2019rkd,
author = {Park, Wonpyo and Kim, Dongju and Lu, Yan and Cho, Minsu},
title = {Relational Knowledge Distillation},
booktitle = {Proceedings of the IEEE/CVF Conference on Computer Vision and Pattern Recognition (CVPR)},
month = {June},
year = {2019}
}

@InProceedings{zhao2022dkd,
    author    = {Zhao, Borui and Cui, Quan and Song, Renjie and Qiu, Yiyu and Liang, Jiajun},
    title     = {Decoupled Knowledge Distillation},
    booktitle = {Proceedings of the IEEE/CVF Conference on Computer Vision and Pattern Recognition (CVPR)},
    month     = {June},
    year      = {2022},
    pages     = {11953-11962}
}

@INPROCEEDINGS{li2023ckd,
  author={Lou, Zhibo and Otake, Shinta and Li, Zhengxiao and Kawakami, Rei and Inoue, Nakamasa},
  booktitle={ICASSP 2024 - 2024 IEEE International Conference on Acoustics, Speech and Signal Processing (ICASSP)}, 
  title={Cubic Knowledge Distillation for Speech Emotion Recognition}, 
  year={2024},
  volume={},
  number={},
  pages={5705-5709}
  }

@INPROCEEDINGS{10887615,
  author={Lin, Hsi-Che and Lin, Yi-Cheng and Chou, Huang-Cheng and Lee, Hung-yi},
  booktitle={ICASSP}, 
  title={Improving Speech Emotion Recognition in Under-Resourced Languages via Speech-to-Speech Translation with Bootstrapping Data Selection}, 
  year={2025},
  volume={},
  number={},
  pages={1-5}
  }

@INPROCEEDINGS{lightsernet,
  author={Aftab, Arya and Morsali, Alireza and Ghaemmaghami, Shahrokh and Champagne, Benoit},
  booktitle={ICASSP}, 
  title={LIGHT-SERNET: A Lightweight Fully Convolutional Neural Network for Speech Emotion Recognition}, 
  year={2022},
  volume={},
  number={},
  pages={6912-6916}
  }

@ARTICLE{10552082,
  author={Chou, Huang-Cheng and Goncalves, Lucas and Leem, Seong-Gyun and Salman, Ali N. and Lee, Chi-Chun and Busso, Carlos},
  journal={IEEE Transactions on Affective Computing}, 
  title={Minority Views Matter: Evaluating Speech Emotion Classifiers With Human Subjective Annotations by an All-Inclusive Aggregation Rule}, 
  year={2025},
  volume={16},
  number={1},
  pages={41-55},
}

@inproceedings{xia2018instance,
  title={Instance weighting for domain adaptation via trading off sample selection bias and variance},
  author={Xia, Rui and Pan, Zhenchun and Xu, Feng},
  booktitle={Proceedings of the 27th International Joint Conference on Artificial Intelligence, Stockholm, Sweden},
  pages={13--19},
  year={2018}
}

@ARTICLE{9745778,
  author={Yang, Guanglei and Fini, Enrico and Xu, Dan and Rota, Paolo and Ding, Mingli and Nabi, Moin and Alameda-Pineda, Xavier and Ricci, Elisa},
  journal={IEEE Transactions on Pattern Analysis and Machine Intelligence}, 
  title={Uncertainty-Aware Contrastive Distillation for Incremental Semantic Segmentation}, 
  year={2023},
  volume={45},
  number={2},
  pages={2567-2581}
  }

@ARTICLE{scholkopf2001estimating,
  author={Schölkopf, Bernhard and Platt, John C. and Shawe-Taylor, John and Smola, Alex J. and Williamson, Robert C.},
  journal={Neural Computation}, 
  title={Estimating the Support of a High-Dimensional Distribution}, 
  year={2001},
}

@INPROCEEDINGS{11434750,
  author={Wei, Jui-Chiang and Lin, Yi-Cheng and Ritter-Gutierrez, Fabian and Lee, Hung-Yi},
  booktitle={2025 IEEE Automatic Speech Recognition and Understanding Workshop (ASRU)}, 
  title={Multi-Distillation from Speech and Music Representation Models}, 
  year={2025},
  volume={},
  number={},
  pages={1-8},
}

@inproceedings{gretton2005hsic,
  title={Measuring statistical dependence with Hilbert-Schmidt norms},
  author={Gretton, Arthur and Bousquet, Olivier and Smola, Alex and Sch{\"o}lkopf, Bernhard},
  booktitle={International conference on algorithmic learning theory},
  pages={63--77},
  year={2005},
  organization={Springer}
}

@inproceedings{romero2015fitnets,
 author = {Srivastava, Rupesh K and Greff, Klaus and Schmidhuber, J\"{u}rgen},
 booktitle = {Advances in Neural Information Processing Systems},
 editor = {C. Cortes and N. Lawrence and D. Lee and M. Sugiyama and R. Garnett},
 pages = {},
 publisher = {Curran Associates, Inc.},
 title = {Training Very Deep Networks},
 volume = {28},
 year = {2015}
}

@InProceedings{tung2019similarity,
    author = {Tung, Frederick and Mori, Greg},
    title = {Similarity-Preserving Knowledge Distillation},
    booktitle = {Proceedings of the IEEE/CVF International Conference on Computer Vision (ICCV)},
    month = {October},
    year = {2019}
}

@InProceedings{passalis2018learning,
    author = {Passalis, Nikolaos and Tefas, Anastasios},
    title = {Learning Deep Representations with Probabilistic Knowledge Transfer},
    booktitle = {Proceedings of the European Conference on Computer Vision (ECCV)},
    year = {2018}
}

@InProceedings{peng2019correlation,
    author = {Peng, Baoyun and Jin, Xiao and Liu, Jiaheng and Li, Dongsheng and Wu, Yichao and Liu, Yu and Zhou, Shunfeng and Zhang, Zhaoning},
    title = {Correlation Congruence for Knowledge Distillation},
    booktitle = {Proceedings of the IEEE/CVF International Conference on Computer Vision (ICCV)},
    year = {2019}
}

@inproceedings{
    tian2020contrastive,
    title={Contrastive Representation Distillation},
    author={Yonglong Tian and Dilip Krishnan and Phillip Isola},
    booktitle={International Conference on Learning Representations},
    year={2020}
}

@inproceedings{zhou2024rethinking,
    author = {Zhou, Zikai and Shen, Yunhang and Shao, Shitong and Gong, Linrui and Lin, Shaohui},
    title = {Rethinking centered kernel alignment in knowledge distillation},
    year = {2024},
    booktitle = {Proceedings of the Thirty-Third International Joint Conference on Artificial Intelligence},
    articleno = {628},
    numpages = {9},
    location = {Jeju, Korea},
    series = {IJCAI '24}
}

@inproceedings{bijoy2025multilingual,
  title     = {{Multi-Teacher Language-Aware Knowledge Distillation for Multilingual Speech Emotion Recognition}},
  author    = {Mehedi Hasan Bijoy and Dejan Porjazovski and Tamás Grósz and Mikko Kurimo},
  year      = {2025},
  booktitle = {{Interspeech 2025}},
  pages     = {146--150},
  doi       = {10.21437/Interspeech.2025-418},
  issn      = {2958-1796},
}

@inproceedings{lin24i_interspeech,
  title     = {{Emo-bias: A Large Scale Evaluation of Social Bias on Speech Emotion Recognition}},
  author    = {Yi-Cheng Lin and Haibin Wu and Huang-Cheng Chou and Chi-Chun Lee and Hung-yi Lee},
  year      = {2024},
  booktitle = {{Interspeech 2024}},
  pages     = {4633--4637},
  doi       = {10.21437/Interspeech.2024-1073},
  issn      = {2958-1796},
}

@inproceedings{
loshchilov2018decoupled,
title={Decoupled Weight Decay Regularization},
author={Ilya Loshchilov and Frank Hutter},
booktitle={International Conference on Learning Representations},
year={2019},
url={https://openreview.net/forum?id=Bkg6RiCqY7},
}

@InProceedings{pmlr-v97-kornblith19a,
  title = 	 {Similarity of Neural Network Representations Revisited},
  author =       {Kornblith, Simon and Norouzi, Mohammad and Lee, Honglak and Hinton, Geoffrey},
  booktitle = 	 {Proceedings of the 36th International Conference on Machine Learning},
  pages = 	 {3519--3529},
  year = 	 {2019},
  editor = 	 {Chaudhuri, Kamalika and Salakhutdinov, Ruslan},
  volume = 	 {97},
  series = 	 {Proceedings of Machine Learning Research},
  publisher =    {PMLR},
}

@inproceedings{lin24b_interspeech,
  title     = {{On the social bias of speech self-supervised models}},
  author    = {Yi-Cheng Lin and Tzu-Quan Lin and Hsi-Che Lin and Andy T. Liu and Hung-yi Lee},
  year      = {2024},
  booktitle = {{Interspeech 2024}},
  pages     = {4638--4642},
  doi       = {10.21437/Interspeech.2024-454},
  issn      = {2958-1796},
}

@misc{lin2026fairspeech,
      title={Toward Fair Speech Technologies: A Comprehensive Survey of Bias and Fairness in Speech AI}, 
      author={Yi-Cheng Lin and Yun-Shao Tsai and Kuan-Yu Chen and Hsiao-Ying Huang and Huang-Cheng Chou and Hung-yi Lee},
      year={2026},
      eprint={2605.01597},
      archivePrefix={arXiv},
      primaryClass={eess.AS},
      url={https://arxiv.org/abs/2605.01597}, 
}
